\documentclass[runningheads]{llncs}
\usepackage[T1]{fontenc}
\usepackage{graphicx}
\usepackage{marvosym}
\usepackage{multirow}
\usepackage{svg}
\usepackage{amsmath}
\usepackage{amsfonts,amssymb}
\usepackage{xcolor}
\usepackage{bm}
\usepackage{booktabs}
\usepackage{colortbl}
\usepackage{gensymb}
\usepackage{hyperref}

\usepackage[ruled,linesnumbered]{algorithm2e}
\hypersetup{
colorlinks=true,
linkcolor=blue,
anchorcolor=black,
citecolor=blue
}

\begin{document}
\title{Enhanced Scale-aware Depth Estimation \\ for Monocular Endoscopic Scenes \\ with Geometric Modeling}
\titlerunning{Enhanced Scale-aware Depth Estimation with Geometric Modeling}
%
\author{Ruofeng Wei\inst{1} \and
Bin Li\inst{2} \and
Kai Chen\inst{1} \and 
Yiyao Ma\inst{1} \and 
Yunhui Liu\inst{2} \and 
Qi Dou\inst{1\textrm{\Letter}}
}
\authorrunning{R. Wei et al.}
%
\institute{Department of Computer Science and Engineering \\ The Chinese University of Hong Kong, Hong Kong, China \and
Department of Mechanical and Automation Engineering, \\ The Chinese University of Hong Kong, Hong Kong, China}
\maketitle              
\begin{abstract}
Scale-aware monocular depth estimation poses a significant challenge in computer-aided endoscopic navigation. 
However, existing depth estimation methods that do not consider the geometric priors struggle to learn the absolute scale from training with monocular endoscopic sequences.
Additionally, conventional methods face difficulties in accurately estimating details on tissue and instruments boundaries.
In this paper, we tackle these problems by proposing a novel enhanced scale-aware framework that only uses monocular images with geometric modeling for depth estimation.
Specifically, we first propose a multi-resolution depth fusion strategy to enhance the quality of monocular depth estimation. To recover the precise scale between relative depth and real-world values, we further calculate the 3D poses of instruments in the endoscopic scenes by algebraic geometry based on the image-only geometric primitives (\emph{i.e.}, boundaries and tip of instruments).
Afterwards, the 3D poses of surgical instruments enable the scale recovery of relative depth maps. By coupling scale factors and relative depth estimation, the scale-aware depth of the monocular endoscopic scenes can be estimated.
We evaluate the pipeline on in-house endoscopic surgery videos and simulated data. The results demonstrate that our method can learn the absolute scale with geometric modeling and accurately estimate scale-aware depth for monocular scenes.
Code is available at: \url{https://github.com/med-air/MonoEndoDepth}

\end{abstract}
\section{Introduction}

Estimating the scale-aware depth from a monocular endoscopic image is a key yet challenging topic in computer-assisted surgery~\cite{yip2023artificial}, especially for next-generation flexible surgical robots which cannot acquire stereo images. It is a prerequisite for down-stream tasks such as 2D-3D image registration, surgical navigation and autonomous tissue manipulation based on modeling of the surgical field. Despite recent progress on the topic~\cite{li2024image,yang2024self}, there remains fundamental problems being unsolved that may prevent its usage in real-world practice. 
First, previous methods struggle with estimating the absolute scale from monocular images. 
Although the evaluation of these methods typically involves re-scaling each estimate using the median ratio between the ground-truth depth and the prediction~\cite{ozyoruk2021endoslam,shao2022self}, it can be challenging to acquire these median ratios in the navigation. 
Second, existing learning-based approaches often trained on relatively small input resolution using lightweight networks cater to real-time application on embedded platforms~\cite{dong2022towards}, this leads to the loss of detailed information being leveraged for more accurate estimation. 

Several scale-aware depth estimation approaches~\cite{wei2024absolute,zhang2022towards} incorporate kinematics data to estimate the real scale from monocular images. For instance, MetricDepthS-Net~\cite{wei2022distilled} utilizes kinematics and camera poses from the ego-motion network~\cite{recasens2021endo} to recover the scale of endoscopic depth estimation. 
DynaDepth~\cite{zhang2022towards} integrates IMU measurements during training to learn accurate scale for monocular depth estimation. 
However, the use of sensors to obtain kinematics information can be costly and their applicability is often limited by specific environments. In addition, some image-based methods~\cite{petrovai2022exploiting,xue2020toward} employ the geometric relationship between the camera and ground from the image to calculate the scale of relative depth, but these methods are mainly applicable in autonomous driving scenarios. For endoscopic scenes, a promising geometric constraint is the 3D modeling of surgical instruments with cylindrical shafts from the image, which can help estimate the scale-aware depth.

To further achieve depth estimates with high boundary accuracy, various techniques have been developed, such as network design improvements~\cite{ranftl2021vision} and the integration of high-level constraints~\cite{lin2023semhint,yang2024depth}.
For example, DPT~\cite{ranftl2021vision} improves depth estimation accuracy by using vision transformers as a backbone. SemHint-MD~\cite{lin2023semhint} leverages semantic segmentation to guide the training of depth network, leading to better depth estimation. Depth Anything~\cite{yang2024depth} which is a depth foundation model refines the depth estimation with an auxiliary segmentation task.
In this work, instead of proposing a new depth estimation method, we demonstrate that existing depth estimation models can be adapted to generate higher-quality results by fusing estimations from multiple resolution inputs. 

In this paper, we propose an image-based pipeline to estimate the scale-aware depth of monocular endoscopic scenes. It utilizes the surgical tools with cylindrical shaft of a known radius as a geometric constraint to recover the absolute scale. In particular, we firstly improve the quality of relative depth estimation based on a multi-resolution depth fusion strategy. Then, we present a geometric-based optimization model to compute the real 3D pose of the instrument from simple geometric primitives (\emph{i.e.}, boundaries and tip of instruments). Based on the 3D pose, the precise scale between relative depth and real world is recovered. Finally, we qualitatively and quantitatively evaluate our pipeline on in-house endoscopic surgery videos and simulated data. Our method exceeds conventional stereo-based approaches, showing the pipeline can achieve depth estimation with real scale on monocular surgical scenes.

\section{Method}

\subsection{Overview of the Scale-aware Depth Estimation Framework}
\begin{figure}[t]
\centering
\includegraphics[width = 1.0\hsize]{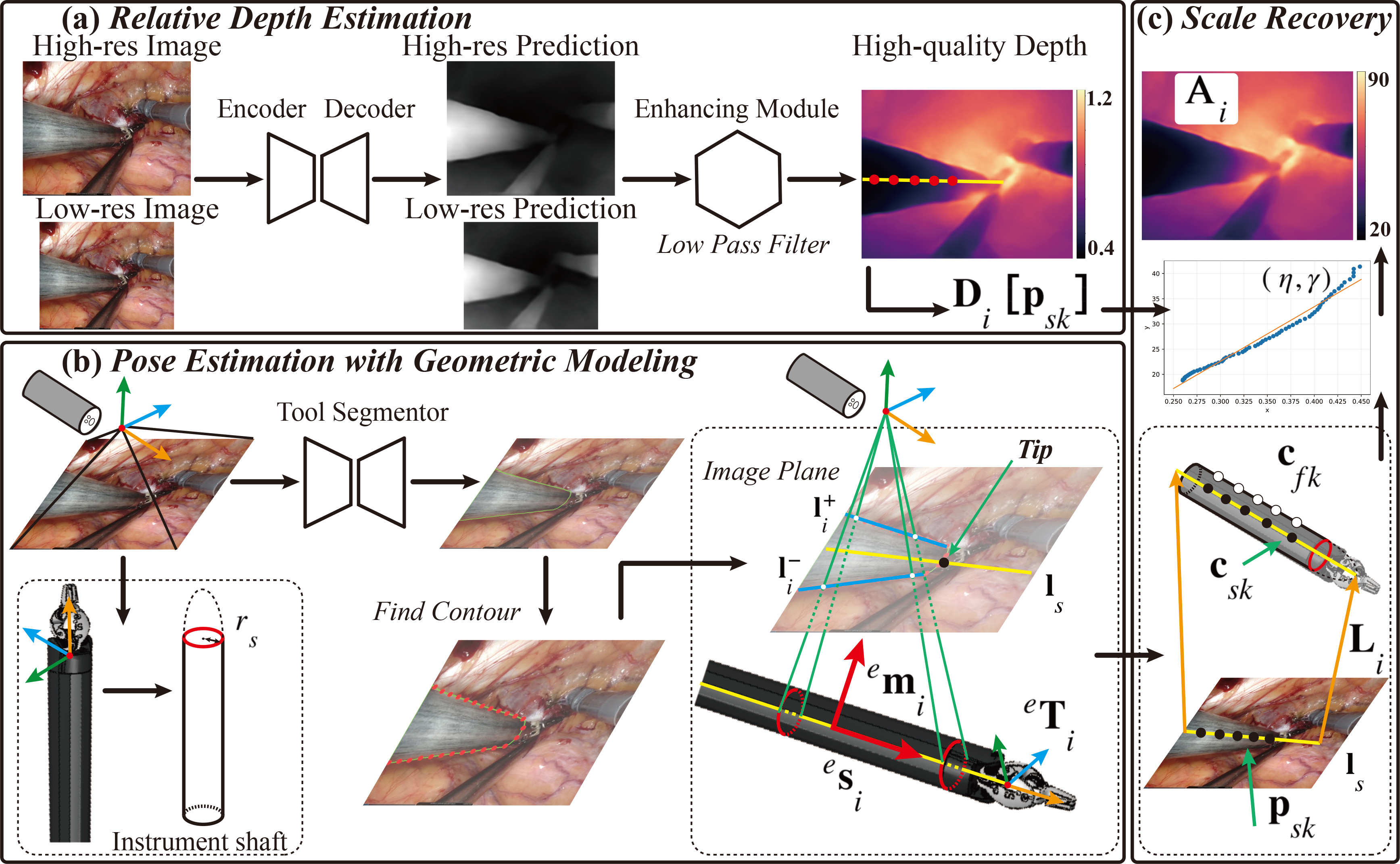}
\caption{Overview of our proposed scale-aware monocular depth estimation framework, which consists of modules for relative depth estimation, surgical instrument pose estimation with geometric modeling, and scale recovery.}
\label{fig: pipeline}
\end{figure}

Fig. \ref{fig: pipeline} shows an overview of our proposed scale-aware monocular depth estimation framework which consists of three parts. First, we fuse estimations at different image resolutions, and use a relative depth estimation network to generate high-quality depth maps. Second, we propose a geometry modeling method to track the poses of the surgical instruments in the image using only geometric primitives from the endoscopic images. Third, the 3D poses of the instruments and the depth map are united to recover the scale between relative depth and real-world values. Thus, the absolute depth of the monocular endoscopic scenes can be estimated via the scale recovery based on surgical instruments geometry.

\subsection{Enhancing Monocular Depth via Multi-Resolution Fusion}

Our relative depth estimation network utilizes Monodepth2~\cite{godard2019digging} as its backbone. In the training phase, the network is learned by minimizing the photo-metric loss between the input images and the corresponding synthetic frames generated through novel view synthesis.
However, since the inputs images are cropped to a relatively low resolution and then fed to the network for training, many details are lost while performing depth estimation. 
To address this issue, we develop an enhancing depth module that fuses two estimates for the same endoscopic image at different resolutions to improve the quality of relative depth maps.

Our depth model is capable of handling arbitrary image sizes for depth estimation. 
As shown in Section A of supplementary material, when input with different resolutions is fed into the network, we can observe a specific trend in the depth maps. In low resolutions that are close to the training data, the estimated depth consistently captures the overall structure but may lack high-frequency details. When the same image is input to the model in higher resolution, more details are captured while the structure consistency of the result gradually degrades. 
Further, the second observation is that when a fusion operation is performed to combine relative depth estimates with different resolutions, the details in the depth estimated from higher-resolution input are transferred to the lower-resolution depth map while maintaining structural consistency. 
Following these observations, we present a module to improve the depth maps in low resolution.

In our enhancing module shown in Fig.~\ref{fig: pipeline} (a), we generate two depth estimations of the same endoscopic image. One depth map estimated from the network with a smaller-res input has a consistent structure but suffers from fine details. Another depth predicted with a higher-res input contains more details. Thus, we adopt a low pass fusing filter~\cite{he2012guided} with low-res depth as a guide and apply it to the high-res depth to generate a higher-quality relative depth with good structural consistency and details. 
Based on the above fusing process, the enhanced relative depth map $\mathbf{D}_i$ for each monocular image is predicted.

\subsection{Pose Estimation of Instruments with Geometric Modeling}
As illustrated in Fig.~\ref{fig: pipeline}(b), most of surgical instruments have a cylindrical shaft with a constant radius $r_s$. Hereby, given an endoscopic frame $i$, we can calculate the 3D pose ${}^e\mathbf{T}_i \in \mathbb{R}^{3\times4}$ of the tool based on its geometrical primitives  (\emph{i.e.}, boundaries and tip), where $e$ denotes the endoscopic image coordinate.
The axis of the tool at pose $^e\mathbf{T}_i$ under the endoscope frame can be represented with the Pl{\"u}cker coordinates $\left( {}^e\mathbf{s}_i, {}^e\mathbf{m}_i\right)$, with ${}^e\mathbf{s}_i \in \mathbb{R}^{3 \times 1}$ being a unit vector denoting the direction of the tool's shaft and ${}^e\mathbf{m}_i \in \mathbb{R}^{3 \times 1}$ being the moment of the tool. Further, assume ${}^e\mathbf{c}_j$ is a 3D point on the tool's surface and ${}^e\widetilde{\mathbf{c}}_j = [{}^e\mathbf{c}_j \; 1]^\mathsf{T}$ is its homogeneous coordinate. Based on the geometrical modeling~\cite{doignon2007degenerate} of cylindrical objects, the following relationship can be derived:
\begin{equation}
    {}^e\widetilde{\mathbf{c}}_j^\mathsf{T} \cdot
    \begin{bmatrix}
        [{}^e\mathbf{s}_i]_\times\,[{}^e\mathbf{s}_i]^\mathsf{T}_\times 
        & [{}^e\mathbf{s}_i]_\times\,{}^e\mathbf{m}_i\\
        {}^e\mathbf{m}_i^\mathsf{T}\,[{}^e\mathbf{s}_i]_\times^\mathsf{T}
        & \|{}^e\mathbf{m}_i\|^2 - r_s^2
    \end{bmatrix}
    \cdot {}^e\widetilde{\mathbf{c}}_j = 0.
\end{equation}

According to the perspective projection theory~\cite{doignon2007degenerate}, the Pl{\"u}cker coordinates of the tool's axis can be associated with the edge boundaries $\left(\mathbf{l}_i^- \in \mathbb{R}^3,  \mathbf{l}_i^+  \in \mathbb{R}^3 \right)$ of the shaft in the image plane as follows:
\begin{equation}
    \mathbf{l}_i^- = \mathbf{K}^{-\mathsf{T}} \cdot \left( \mathbf{I} - \alpha [{}^e\mathbf{s}_i]_\times\right) {}^e\mathbf{m}_i, \quad \mathbf{l}_i^+ = \mathbf{K}^{-\mathsf{T}} \cdot \left( \mathbf{I} + \alpha [{}^e\mathbf{s}_i]_\times\right) {}^e\mathbf{m}_i. 
    \label{eq:edge_line}
\end{equation}
with $\mathbf{K}$ is the camera intrinsic and $\alpha = \frac{r_s}{\sqrt{\|{}^e\mathbf{m}_i\|^2 - r_s^2}}$. Please refer to Section B of supplementary material for detailed derivation. Besides, based on the shaft mask predicted by the tool segmentor, we can compute the edge boundaries from the mask contour. After that, the $\left( {}^e\mathbf{s}_i, {}^e\mathbf{m}_i\right)$ can be calculated via combining Eq.~\ref{eq:edge_line} and the boundaries equations from the 2D mask. 
In this regard, we can obtain the 3D pose ${}^e\mathbf{T}_i$ of the surgical instrument from the Pl{\"u}cker coordinates $[{}^e\mathbf{s}_i,{}^e\mathbf{m}_i]$.

\subsection{Scale Recovery by 3D Poses}

In this section, we first calculate the 3D points on the surface of the shaft on the basis of the calculated pose. By jointly taking these points and corresponding depth information from the estimated relative depth map in Section 2.2, we can compute the transformation scale parameters between the relative depth and the real world.
First, the $\left({}^e\mathbf{s}_i,{}^e\mathbf{m}_i\right)$ can be further transformed to the Pl{\"u}cker matrix $\mathbf{L}_i$ for the representation of the axis:
\begin{equation}
    \mathbf{L}_i = 
    \begin{bmatrix}
        \lbrack {}^e\mathbf{m}_i \rbrack_\times & {}^e\mathbf{s}_i \\
        -{}^e\mathbf{s}_i^\mathsf{T} & 0
    \end{bmatrix}.
\end{equation}
The projection of the axis of the tool in the image plane is calculated as $\mathbf{l}_s = \mathbf{K}^{-\mathsf{T}} \cdot {}^e\mathbf{m}_i$. As shown in Fig.~\ref{fig: pipeline} (b), the shaft tip $\mathbf{p}_{s0} = \lbrack u_{s0} \; v_{s0} \; 1\rbrack^\mathsf{T}$ along $\mathbf{l}_s$ is firstly predicted from the endoscopic image, where $(u_{s0},v_{s0})$ denotes the image pixel. 
The corresponding 3D point $\mathbf{c}_{s0}$ along the axis of shaft is developed as follows:
\begin{equation}
\begin{aligned}
        & \mathbf{w} = \lbrack \mathbf{K} | \mathbf{0}_{3 \times 1} \rbrack \cdot \mathbf{L}_i \cdot \lbrack \mathbf{K}[0][0] \ 0 \ \mathbf{K}[0][2]\!-\!u_{s0} \ 0 \rbrack^\mathsf{T}, \\
        & \to \ \mathbf{c}_{s0} = \lbrack \frac{\mathbf{w}[0]}{\mathbf{w}[3]} \ \frac{\mathbf{w}[1]}{\mathbf{w}[3]} \ \frac{\mathbf{w}[2]}{\mathbf{w}[3]} \rbrack.
\end{aligned}
\end{equation}
where $ \mathbf{w}$ is an vector calculated by $\mathbf{L}_i$ and camera intrinsic.
So the depth of the point $\mathbf{c}_{f0}$ on the surface of the shaft is derived (see Section C of supplementary material). Based on the above process, the depth values $\left(z(\mathbf{c}_{f0}), \cdots, z(\mathbf{c}_{fn})\right)$ of 3D points re-projected by the pixel $\left(\mathbf{p}_{s0}, \cdots, \mathbf{p}_{sn}\right)$ along $\mathbf{l}_s$ can be computed.

Then, we obtain the relative depth values of $\left(\mathbf{p}_{s0},\cdots,\mathbf{p}_{sn}\right)$ from the depth map $\mathbf{D}_i$, which are defined as $\left(\mathbf{D}_i[\mathbf{p}_{s0}], \cdots, \mathbf{D}_i[\mathbf{p}_{sn}]\right)$. Therefore, the transformation scale parameters between the relative depth and the real world can be solved in closed form as a standard least-squares problem:
\begin{equation}
    \left(\eta, \gamma\right) = \mathop{\arg\min}\limits_{\eta, \gamma}\sum_{k=1}^{n}{\left( \eta \times z(\mathbf{c}_{fk}) + \gamma - \mathbf{D}_i[\mathbf{p}_{sk}] \right)}.
\end{equation}
Afterward, as shown in Fig.~\ref{fig: pipeline} (c), we can recover the absolute depth $\mathbf{A}_i$ of the monocular endoscopic image utilizing the parameters $\left(\eta, \gamma\right)$.

\section{Experiments}

{\textbf{Datasets.}} We evaluate the accuracy of the proposed scale-aware depth estimation pipeline on our in-house surgery and simulator data. We extract a total of 95 clips from 6 surgical videos of our in-house daVinci robotic prostatectomy data. The monocular endoscopic sequences in these clips are used for training and evaluation. The instruments used in surgery are standard daVinci instruments with a cylindrical shaft with a constant radius 4.5mm. For training, validation, and testing, there are 3947, 1189, and 534 frames, respectively. In the experiment, the frames are resized from the resolution $1280 \times 1024$ to $640 \times 512$. We also calculate the ground truth depth maps for each frame in our in-house data via standard stereo matching~\cite{wei2022stereo}. 
Furthermore, we collect several video clips from 5 tasks in SurRoL~\cite{xu2021surrol} surgical simulator for quantitative 3D pose evaluation of the scale-aware depth estimation.  

\noindent {\textbf{Evaluation Metrics.}} We report the difference between the predicted and ground truth depth maps using six popular depth metrics~\cite{eigen2014depth}: absolute relative error (Abs Rel), squared relative error (Sq Rel), root mean squared error (RMSE), log-scale RMSE ($\text{RMSE}_{log}$), $\delta<1.25$ which denotes percentage of pixels within 20\% of ground truth values, and mean absolute error (MAE).

\noindent {\textbf{Implementation Details.}} We follow the training process of Monodepth2~\cite{godard2019digging} to learn the relative depth estimation, including pre-trained in ImageNet and trained in our medical data. 
For tool segmentation, we employ a lightweight U-Net with VGG11~\cite{simonyan2014very} as the backbone. The network consists of five scales of down-sample layers to yield a runtime of approximately 12$ms$ per frame. To train the segmentation model, we utilize a publicly available dataset~\cite{allan20192017}, and then apply it to predict binary tool masks on our surgical datasets. To address the domain gap between the training and testing data, we apply morphological operation to refine the tool boundaries in the masks.

\begin{table}[t]
\caption{Quantitative comparisons for scale-aware depth estimation on in-house data. Sq Rel, RMSE, and MAE are in mm. The closer the scale is to 1, the better.  The best results are indicated in bold.}
\label{table: quantitative_depth}
\centering
\resizebox{1.0\hsize}{!}{\begin{tabular}{c|c|ccccc|c}
\specialrule{0.12em}{0pt}{2pt}
\multirow{2}{*}{\text{Method}} & \multirow{2}{*}{\text{Scale}} & \multicolumn{5}{c|}{$\text{Error} \downarrow$}       & \multicolumn{1}{c}{$\text{Accuracy} \uparrow$} \\ \cline{3-8} 
                        &                        & \text{Abs Rel} & \text{Sq Rel} & \text{RMSE}  & $\text{RMSE}_{log}$ & \text{MAE} & $\delta\!<\!1.25$ \\ \specialrule{0.02em}{2pt}{2pt}
EndoSfM\cite{ozyoruk2021endoslam} & NA & 0.165\scriptsize$\pm$0.057  & 2.481\scriptsize$\pm$1.723 & 11.032\scriptsize$\pm$3.646 & 0.200\scriptsize$\pm$0.057  & 8.564\scriptsize$\pm$2.804 & 0.769\scriptsize$\pm$0.115 \\ 
AF-SfMLearner\cite{shao2022self} & NA & 0.211\scriptsize$\pm$0.086  & 4.432\scriptsize$\pm$3.731 & 13.435\scriptsize$\pm$4.873 & 0.250\scriptsize$\pm$0.085   & 10.342\scriptsize$\pm$3.799         & 0.725\scriptsize$\pm$0.118 \\
ManyDepth\cite{watson2021temporal} & NA & 0.165\scriptsize$\pm$0.047  & 2.489\scriptsize$\pm$1.526 & 11.691\scriptsize$\pm$3.599 & 0.204\scriptsize$\pm$0.052   & 9.608\scriptsize$\pm$2.717 & 0.742\scriptsize$\pm$0.110        \\ 
Depth Anything\cite{yang2024depth} & NA  & 0.179\scriptsize$\pm$0.053  & 3.734\scriptsize$\pm$2.272 & 15.371\scriptsize$\pm$5.206 & 0.220\scriptsize$\pm$0.062   & 11.177\scriptsize$\pm$3.747 & 0.710\scriptsize$\pm$0.134        \\
DPT\cite{ranftl2021vision} & NA  & 0.180\scriptsize$\pm$0.060  & 3.201\scriptsize$\pm$2.876 & 13.024\scriptsize$\pm$4.952 & 0.222\scriptsize$\pm$0.062   & 9.997\scriptsize$\pm$3.584 & 0.719\scriptsize$\pm$0.128        \\
MonoDepth Stereo\cite{recasens2021endo} & 1.197\scriptsize$\pm$0.147  & 0.182\scriptsize$\pm$0.052  & 3.078\scriptsize$\pm$1.668 & 13.194\scriptsize$\pm$3.785 & 0.226\scriptsize$\pm$0.056   & 10.285\scriptsize$\pm$2.904 & 0.699\scriptsize$\pm$0.129        \\ \specialrule{0.02em}{2pt}{2pt}
Ours               & \bf0.959\scriptsize$\pm$0.043            & \bf0.110\scriptsize$\pm$0.043  & \bf1.388\scriptsize$\pm$1.340 & \textbf{8.637}\scriptsize$\pm$3.958 & \textbf{0.148}\scriptsize$\pm$0.053   & \textbf{6.347}\scriptsize$\pm$2.943         & \textbf{0.880}\scriptsize$\pm$0.102        \\ \specialrule{0.12em}{2pt}{0pt}
\end{tabular}}
\end{table}

\subsection{Results}

\begin{figure}[t]
\centering
\includegraphics[width = 1.0\hsize]{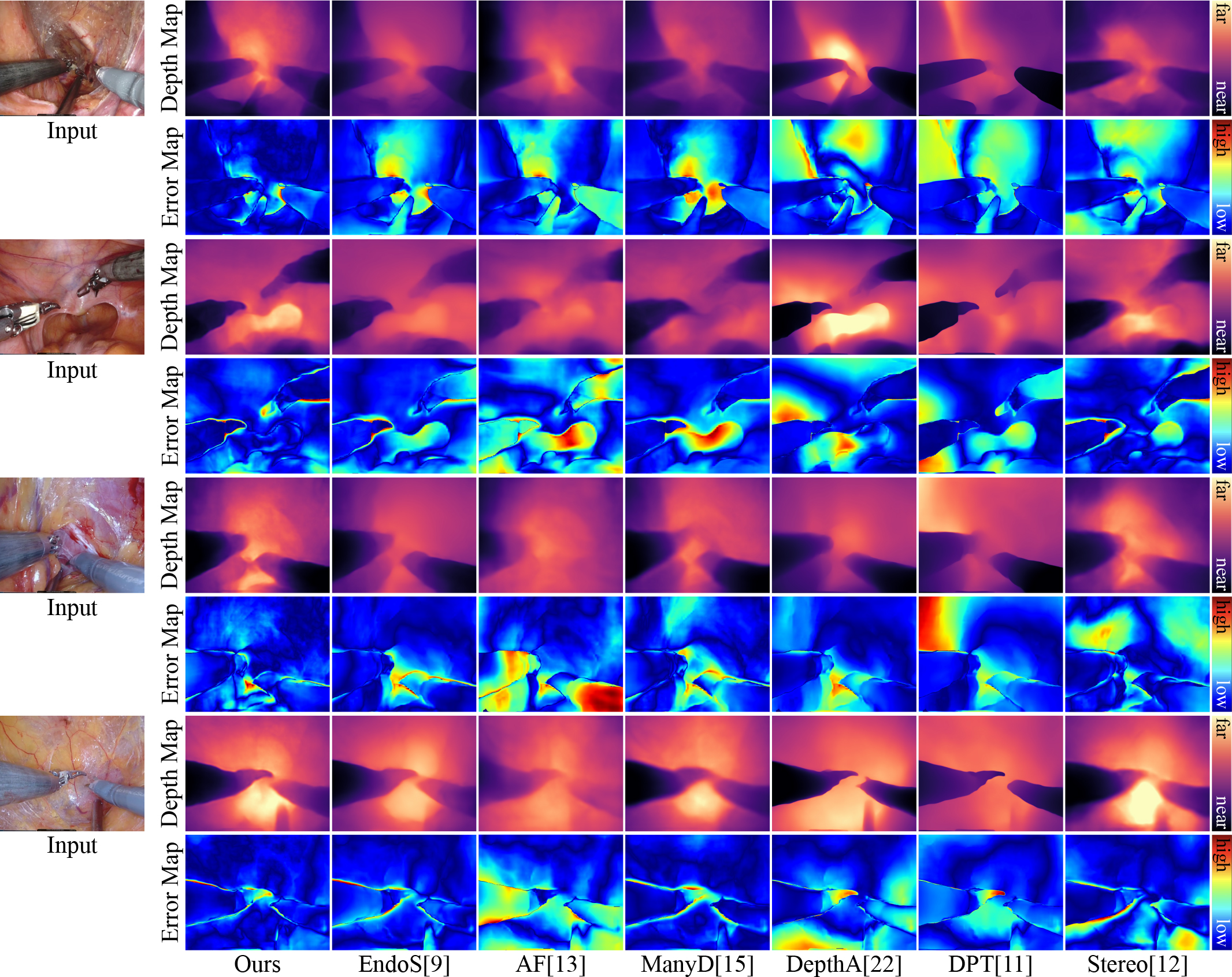}
\caption{Qualitative comparisons on in-house data. Our method outperforms EndoSfM(EndoS)~\cite{ozyoruk2021endoslam}, AF-SfMLearner(AF)~\cite{shao2022self}, ManyDepth(ManyD)~\cite{watson2021temporal}, Depth Anything(DepthA)~\cite{yang2024depth}, DPT~\cite{ranftl2021vision}, and MonoDepth Stereo(Stereo)~\cite{recasens2021endo} in depth quality.}
\label{fig: qualitative_depth}
\end{figure}

\textbf{Evaluation on Scale-Aware Depth Estimation.} We present the quantitative depth comparison results on our in-house surgery data in Table~\ref{table: quantitative_depth}, which rescale the results using the ground truth median scaling method. Apart from depth metrics, we calculate the means and standard errors of the re-scaling factors to showcase the scale-awareness capability. Our proposed depth model demonstrates the best performance in terms of up-to-scale accuracy across all metrics. Specifically, the state-of-the-art baseline method only achieves an MAE error of 8.564 mm, an RMSE of 11.032 mm, and an accuracy of 76.9\%. In contrast, our method demonstrates superior performance with an MAE error of 6.347 mm, an RMSE of 8.637 mm, and an accuracy of 88.0\%. Notably, our model also achieves near-perfect accuracy in terms of absolute scale.  These quantitative results show that the proposed method indeed extracts the absolute scale with geometric modeling, resulting in fine scale-aware depth estimation. Furthermore, four typical images are selected for qualitative depth comparison. As shown in Fig.~\ref{fig: qualitative_depth}, our method can predict the scale-aware depth with smaller errors, sharp boundaries and fine-grained details compared to other approaches. Moreover, the runtime of the proposed method is around 66 ms per frame.

\begin{figure}[t]
\centering
\includegraphics[width = 1.0\hsize]{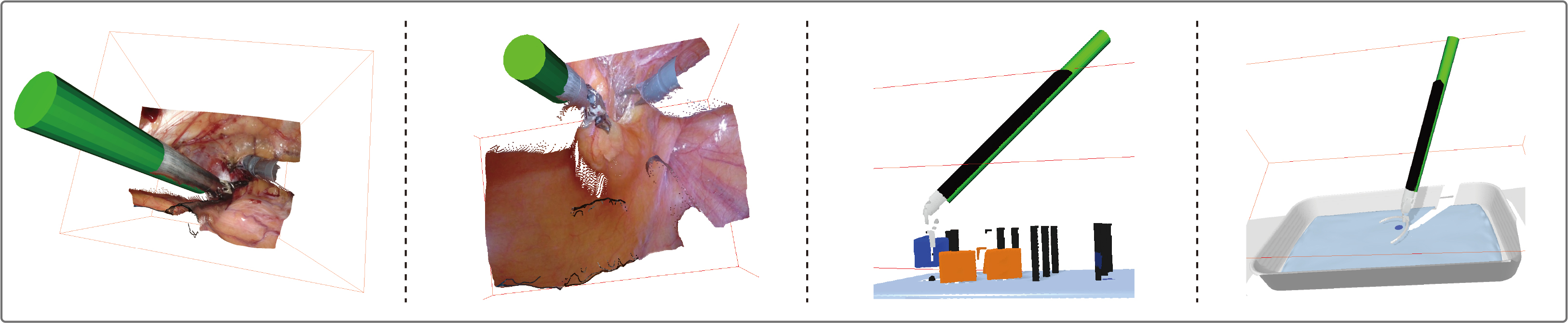}
\caption{Qualitative comparison of 3D pose estimation. Green cylinders represent the rendered calculated poses of surgical tools.}
\label{fig: qualitative_pose}
\end{figure}

\noindent \textbf{Evaluation on Instrument Pose Estimation.} We present the quantitative pose evaluation results on simulator data in Table~\ref{tab:pose}. In addition, as illustrated in Fig.~\ref{fig: qualitative_pose}, the calculated poses of the surgical instruments are rendered and put together with ground truth depth maps for qualitative comparison. Table~\ref{tab:pose} shows that the average pose estimation errors in orientation and translation are 1.459\,\degree, 3.220\,\degree, 1.220\,mm, 1.150\,mm, and 2.356\,mm, indicating that our geometric modeling-based method can predict the tool's poses with high accuracy. Besides, the qualitative comparison results demonstrate the calculated 3D poses of the instruments are aligned well with the ground truth depth, thereby proving the effectiveness of the proposed pose estimation method.

\begin{table}[t]
\caption{Quantitative evaluation of 3D pose estimation of surgical instruments.The orientation errors are measured in degrees, while the localization errors are in mm.}
\resizebox{1.0\hsize}{!}{
\begin{tabular}{c|c|ccccc}
\specialrule{0.12em}{0pt}{2pt}
Task                & Im. \# & Ori. X & Ori. Y & Tip X & Tip Y & Tip Z \\ \specialrule{0.02em}{2pt}{2pt}
Peg Transfer        & 49     &  1.249\scriptsize$\pm$1.291      & 2.909\scriptsize$\pm$3.053       & 1.076\scriptsize$\pm$0.676      & 1.141\scriptsize$\pm$0.759      &  2.415\scriptsize$\pm$1.787      \\
Needle Pick         & 36       & 1.312\scriptsize$\pm$1.274       & 2.877\scriptsize$\pm$3.065       & 1.342\scriptsize$\pm$0.672      & 1.127\scriptsize$\pm$0.727      & 2.484\scriptsize$\pm$2.322      \\
Needle Reach        & 42       &  1.503\scriptsize$\pm$1.380      & 3.367\scriptsize$\pm$3.243       & 1.121\scriptsize$\pm$0.752      & 1.105\scriptsize$\pm$0.728      & 2.241\scriptsize$\pm$1.835      \\
Gauze Retrieve      & 43       &  1.723\scriptsize$\pm$1.525      & 3.680\scriptsize$\pm$2.954       & 1.273\scriptsize$\pm$0.757      & 1.178\scriptsize$\pm$0.812      & 2.325\scriptsize$\pm$2.415      \\
Bimual Peg Transfer & 37       & 1.507\scriptsize$\pm$1.617       & 3.267\scriptsize$\pm$2.841       & 1.290\scriptsize$\pm$0.697      & 1.198\scriptsize$\pm$0.863      & 2.316\scriptsize$\pm$2.075      \\ \specialrule{0.02em}{2pt}{2pt}
Mean                &  \bf 41     &   \bf1.459     &  \bf3.220      &  \bf1.220     &  \bf1.150     & \bf2.356      \\ \specialrule{0.12em}{2pt}{0pt}
\end{tabular}
}
\label{tab:pose}
\end{table}

\noindent \textbf{Ablation Studies.} To study the impact of different input resolution in the enhancing module, we perform a quantitative ablation on the enhancing module of our relative depth estimation network in Table~\ref{tab:ablation}. We observe that the multi-resolution fusion strategy indeed improves the quality of relative depth estimation. Furthermore, when the high-res input images are set to $480\times384$, the relative depth network can obtain higher depth estimation accuracy.

\begin{table}[!h]
\caption{Ablation studies of different input resolution in the enhancing module.}
\label{tab:ablation}
\centering
\begin{tabular}{cc|c|ccc|c}
\specialrule{0.12em}{0pt}{2pt}
\multicolumn{2}{c|}{\text{High-res}} & \multicolumn{1}{c|}{\text{Low-res}} & \multicolumn{3}{c|}{$\text{Error} \downarrow$}       & \multicolumn{1}{c}{$\text{Accuracy} \uparrow$} \\ \cline{1-7} \specialrule{0.02em}{0pt}{2pt}
   $640 \times 512$ & $480 \times 384$ &  $320 \times 256$ & \text{Abs Rel} & \text{RMSE}  & $\text{MAE}$  & $\delta\!<\!1.25$ \\ \specialrule{0.02em}{2pt}{2pt}
 & & \checkmark  & 0.113\scriptsize$\pm$0.045 & 9.008\scriptsize$\pm$3.038 & 6.524\scriptsize$\pm$2.068 & 0.878\scriptsize$\pm$0.073 \\ 
\checkmark &  & \checkmark  & 0.113\scriptsize$\pm$0.045 & 8.881\scriptsize$\pm$3.028 & 6.424\scriptsize$\pm$2.068  & 0.881\scriptsize$\pm$0.072\\
 & \checkmark & \checkmark  & \textbf{0.110}\scriptsize$\pm$0.045 & \textbf{8.698}\scriptsize$\pm$3.065 & \textbf{6.322}\scriptsize$\pm$2.091 & \textbf{0.886}\scriptsize$\pm$0.074   \\ \specialrule{0.12em}{2pt}{0pt}
\end{tabular}
\end{table}

\section{Conclusion}

This paper presents a novel image-based method for high-quality and scale-aware monocular depth estimation with geometry modeling. 
We first improve the relative depth via a fusion strategy. Then, we compute the instruments poses using algebraic geometry. Leveraging these 3D poses, the accurate scale is recovered, enabling scale-aware depth estimation of monocular images. 
We also evaluate the accuracy of the proposed method on our surgical and simulated data.
In the future, the scale-aware depth estimation will be utilized in robotic ENT surgery.

\begin{credits}
\subsubsection{\ackname} This work was supported in part by the Shenzhen Portion of Shenzhen-Hong Kong Science and Technology Innovation Cooperation Zone under HZQB-KCZYB-20200089, in part by the National Natural Science Foundation of China under Project No. 62322318, in part by the ANR/RGC Joint Research Scheme of the Research Grants Council of the Hong Kong Special Administrative Region, China and the French National Research Agency (Project No. A-CUHK402/23), and in part by Hong Kong Innovation and Technology Commission under Project No. PRP/026/22FX.

\subsubsection{\discintname}
The authors have no competing interests to declare that are relevant to the content of this article.

\end{credits}
%
%
%
%

\bibliographystyle{splncs04}
\bibliography{Paper-1856}

\end{document}